\documentclass{isprs}
\usepackage{setspace}
\usepackage{geometry} % added 27-02-2014 Markus Englich
\usepackage{graphicx}
\usepackage{subfig}
\usepackage{fontenc}
\usepackage{color}
\usepackage[svgnames]{xcolor}
\usepackage{amsmath}
\usepackage{url}
\usepackage{pgfplots}
\usepackage{booktabs}

\pgfplotsset{
  compat=1.9,
  xlabel near ticks,
  ylabel near ticks
}

\begin{document}

\title{semantic3D.net: A new Large-scale Point Cloud Classification Benchmark}

% KAO: Remove extra spacing
\author{
 Timo Hackel\textsuperscript{a}, Nikolay Savinov\textsuperscript{b}, Lubor Ladicky\textsuperscript{b}, Jan D. Wegner\textsuperscript{a}, Konrad Schindler\textsuperscript{a}, Marc 
Pollefeys\textsuperscript{b}}

% KAO: Remove extra newline
\address{
	\textsuperscript{a }IGP, ETH Zurich, Switzerland - (timo.hackel, jan.wegner, konrad.schindler)@geod.baug.ethz.ch\\
	\textsuperscript{b }CVG, ETH Zurich, Switzerland - (nikolay.savinov, lubor.ladicky, marc.pollefeys)@inf.ethz.ch\\
}

%\author{Timo Hackel, Nikolay Savinov, Lubor Ladicky, Jan D. Wegner, Konrad Schindler and Marc Pollefeys\\
%ETH Zurich, Switzerland\\
%{\tt\small \{timo.hackel@geod.baug, nikolay.savinov@inf, lubor.ladicky@inf,}\\
%{\tt\small jan.wegner@geod.baug, konrad.schindler@geod.baug,marc.pollefeys@inf\}.ethz.ch}
%}

%\address
%{
  %Photogrammetry and Remote Sensing, ETH Z\"urich --
  %firstname.lastname@geod.baug.ethz.ch
%}

\commission{II, }{II} %This field is optional.
\workinggroup{II/6} %This field is optional.

%\definecolor{bblue}{HTML}{4F81BD}
%\definecolor{rred}{HTML}{C0504D}
%
%\hyphenation{data-bases}
%\def\hyph{-\penalty0\hskip0pt\relax}

\abstract{
\vspace{-0.5em}

This paper presents a new 3D point cloud classification benchmark data
set with over four billion manually labelled points, meant as input for
data-hungry (deep) learning methods.
We also discuss first submissions to the benchmark that use deep
convolutional neural networks (CNNs) as a work horse, which already
show remarkable performance improvements over state-of-the-art.
CNNs have become the de-facto standard for many tasks in computer
vision and machine learning like semantic segmentation or object
detection in images, but have no yet led to a true breakthrough for 3D
point cloud labelling tasks due to lack of training data.
With the massive data set presented in this paper, we aim at closing
this data gap to help unleash the full potential of deep learning
methods for 3D labelling tasks.
%
%Recent advances in Machine Learning and Computer Vision have shown that complex 
%real-world classification tasks require large amounts of data for training. Yet, 
%producing such amounts for 3D points requires substantial effort and well-designed 
%labelling interface.
%
Our \textit{semantic3D.net} data set consists of dense point clouds
acquired with static terrestrial laser scanners. It contains $8$
semantic classes and covers a wide range of urban outdoor scenes:
churches, streets, railroad tracks, squares, villages, soccer fields
and castles.
We describe our labelling interface and show that our data set
provides more dense and complete point clouds with much higher overall
number of labelled points compared to those already available to the
research community. We further provide baseline method descriptions
and comparison between methods submitted to our online system.
We hope \textit{semantic3D.net} will pave the way for deep learning
methods in 3D point cloud labelling to learn richer, more general 3D
representations, and first submissions after only a few months
indicate that this might indeed be the case.  }

\maketitle

\section{Introduction}\label{sec:introduction}
\vspace{-0.5em}

%The market for sensors has shown an enormous growth in the past decades and is worth 
%multiple billion dollar each year. Nowadays even simple products contain a multitude 
%of sensors. 
%%
%However, with many sensors comes big data. What makes big data challenging is 
%not only the data itself but also the tasks to be performed. Typically, such tasks 
%perform best, when they are data driven. 
%%
%Especially, perceptual systems have shown significant improvements given large data sets, 
%large computational resources and powerful inference techniques, such as convolutional 
%neural networks.

Deep learning has made a spectacular comeback since the seminal paper
of \cite{krizhevsky2012imagenet}, which revives earlier work of
\cite{fukushima1980neocognitron,lecun1989backpropagation}. Especially
deep convolutional neural networks (CNN) have quickly become the core
technique for a whole range of learning-based image analysis
tasks. The large majority of state-of-the-art methods in computer
vision and machine learning include CNNs as one of their essential
components.
Their success for image-interpretation tasks is mainly due to (i)
easily parallelisable network architectures that facilitate training
from millions of images on a single GPU and (ii) the availability of
huge public benchmark data sets like
\textit{ImageNet}~\cite{deng2009imagenet,russakovsky2015} and
\textit{Pascal VOC}~\cite{everingham2010pascal} for rgb images, or
\textit{SUN rgb-d}~\cite{song2015sun} for rgb-d data.

\begin{figure}[htbp]
  \centering
  \includegraphics[width=\columnwidth]{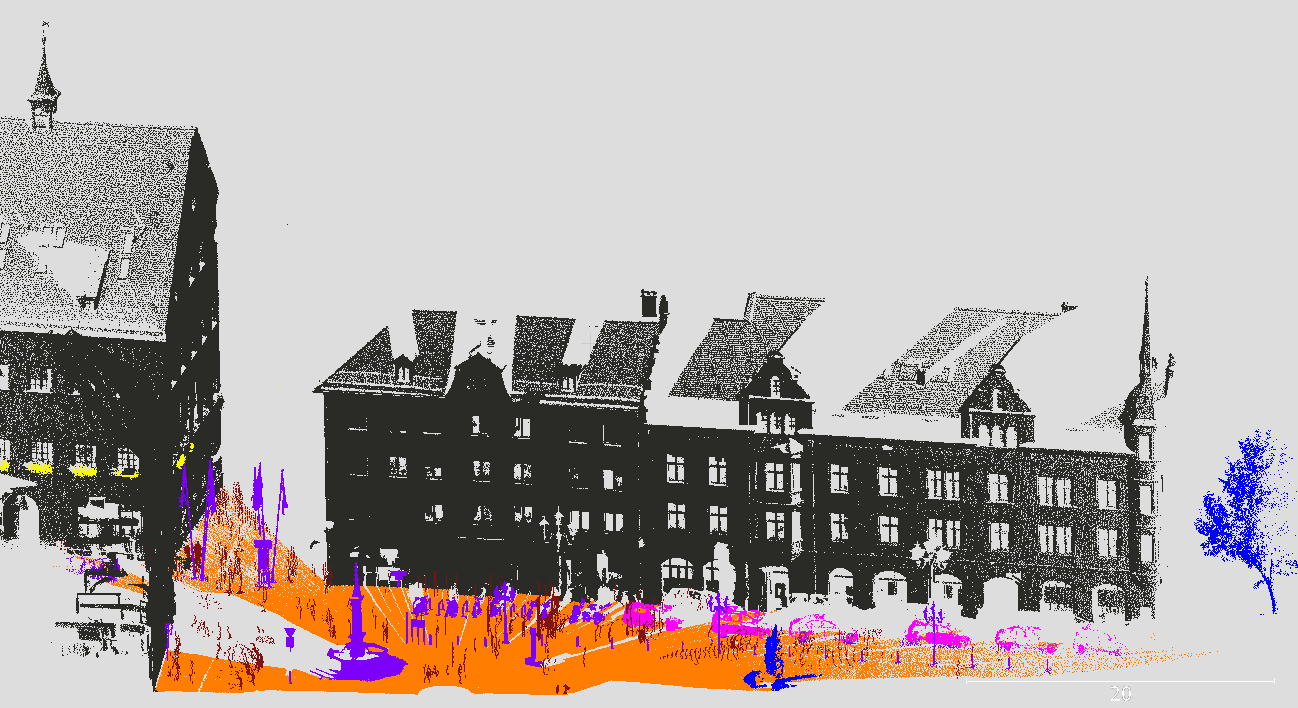}
  \caption{Example point cloud of the benchmark dataset, where colours indicate class labels.}
  \label{fig:intr}
\end{figure}

While CNNs have been a great success story for image interpretation,
it has been less so for 3D point cloud interpretation. 
%Transforming 3D point clouds to depth images is usually not the best option, because
%it results in significant loss of information depending on the
%scanning
%process~\cite{blug2004fast,pfotzer2014development}. \textcolor{red}{
  %slightly unqualified argument, depends on dataset. For single polar
  %scans (like in our benchmark) the loss is 0, they have been recorded
  %as panoramic depth images} 
%This makes the development of computer vision algorithms for point clouds even more challenging.
%
%Yet, imaging is only one in many technologies used for perceptive sensing and each of 
%these technologies comes with their own advantages and disadvantages. Another widespread 
%technology is laser scanning. 
%
%Certainly, one of the biggest strengths of laser scanning is the ability to obtain 
%accurate and dense 3D point clouds, which capture the geometry of scenes with a high 
%density, little noise and that directly deliver metric scale (see Figure \ref{fig:labeling}). 
%
%
What makes supervised learning hard for 3D point clouds is the sheer
size of millions of points per data set, and the irregular, not
grid-aligned, and in places very sparse structure, with strongly
varying point density (Figure~\ref{fig:intr}).

While recording is nowadays straight-forward, the main bottleneck is
to generate enough manually labeled training data, needed for
contemporary (deep) machine learning to learn good models, which
generalize well across new, unseen scenes.
Due to the additional dimension, the number of classifier parameters
is larger in 3D space than in 2D, and specific 3D effects like
occlusion or variations in point density lead to many different
patterns for identical output classes. This aggravates training good,
general classifiers and we generally need more training data in 3D
than in 2D\footnote{Note that the large number of 3D points of
  \textit{semantic3d.net} ($4 \times 10^{9}$ points) is at the same
  scale as the number of pixels of the \textit{SUN rgb-d} benchmark
  ($\approx 3.3 \times 10^{9}$ px)~\cite{song2015sun}, which aims at
  3D object classification. However, the number of 3D points per laser
  scan ($\approx 4 \times 10^{8}$ points) is much larger than the
  number of pixels per image ($\approx 4 \times 10^{5}$ px).}.
In contrast to images, which are fairly easy to annotate even for
untrained users, 3D point clouds are harder to interpret. Navigation
in 3D is more time-consuming and the strongly varying point density
aggravates scene interpretation.

In order to accelerate the development of powerful algorithms for
point cloud processing\footnote{Note that, besides laser scanner point clouds, it is 
also more efficient to classify point clouds generated from 
SfM pipelines directly instead of going through all individual images to then merge results
\cite{riemenschneider2014learning}.}, we provide the (to our knowledge) hitherto
largest collection of (terrestrial) laser scans with point-level
semantic ground truth annotation. In total, it consists of over $4
\cdot 10^9$ points and class labels for $8$ classes. The data set is
split into training and test sets of approximately equal size.
The scans are challenging, not only due to their size of up to
$\approx 4 \cdot 10^8$ points per scan, but also because of their high
measurement resolution and long measurement range, leading to extreme
density changes and large occlusions.
For convenient use of the benchmark test we provide not only freely
available data, but also an automated online submission system as well
as public results of the submitted methods.
The benchmark also includes baselines, one following the standard
paradigm of eigenvalue-based feature extraction at multiple scales
followed by classification with a random forest, the other a basic
deep learning approach. Moreover, first submissions have been made to
the benchmark, which we also briefly discuss.

\section{Related Work}\label{sec:related_work}
\vspace{-0.5em}

Benchmarking efforts have a long tradition in the geospatial data
community and particularly in ISPRS. Recent efforts include, for
example, the \textit{ISPRS-EuroSDR benchmark on High Density Aerial
  Image
  Matching}\footnote{\url{http://www.ifp.uni-stuttgart.de/ISPRS-EuroSDR/ImageMatching/index.en.html}}
that aims at evaluating dense matching methods for oblique aerial
images \cite{haala2013,cavegn2014} and the \textit{ISPRS Benchmark
  Test on Urban Object Detection and Reconstruction}, which contains
several different challenges like semantic segmentation of aerial
images and 3D object reconstruction \cite{rottensteiner2013}.

In computer vision, very large-scale benchmark data sets with millions
of images have become standard for learning-based image interpretation
tasks. A variety of datasets have been introduced, many tailored for a
specific task, some serving as basis for annual challenges for several
consecutive years (e.g., \textit{ImageNet, Pascal VOC}).
Datasets that aim at boosting research in image classification and
object detection heavily rely on images downloaded from the
internet. Web-based imagery has been a major driver of benchmarks
because no expensive, dedicated photography campaigns have to be
accomplished for dataset generation. This enables scaling benchmarks
from hundreds to millions of images, although often weakly annotated
and with a considerable amount of label noise that has to be taken
into account.  Additionally, one can assume that internet images
constitute a very general collection of images with less bias towards
particular sensors, scenes, countries, objects etc., which allows
training richer models that generalize well.

One of the first successful attempts to object detection in images at
very large scale is \textit{tinyimages} with over 80 million small
($32 \times 32~px$) images \cite{torralba200880}.
A milestone and still widely used dataset for semantic image
segmentation is the famous Pascal VOC \cite{everingham2010pascal}
dataset and challenge, which has been used for training and testing
many of the well-known, state-of-the-art algorithms today
like~\cite{long2015fully,badrinarayanan2015segnet}.  Another, more
recent dataset is \textit{MSCOCO}\footnote{\url{http://mscoco.org/}},
which contains $300{,}000$ images with annotations that allow for object
segmentation, object recognition in context, and image captioning.
One of the most popular benchmarks in computer vision today is the
\textit{ImageNet} dataset~\cite{deng2009imagenet,russakovsky2015},
which made Convolutional Neural Networks popular in computer
vision~\cite{krizhevsky2012imagenet}. It contains $ >14\times 10^{6}$
images organized according to the WordNet
hierarchy\footnote{\url{https://wordnet.princeton.edu/}}, where words
are grouped into sets of cognitive synonyms.

The introduction of the popular, low-cost gaming device Microsoft
Kinect gave rise to several, large rgb-d image databases.  Popular
examples are the \textit{NYU Depth Dataset
  V2}~\cite{silberman2012indoor} or \textit{SUN RGB-D}
\cite{song2015sun} that provide labeled rgb-d images for object
segmentation and scene understanding. Compared to laser scanners,
low-cost, structured-light rgb-d sensors have much shorter measurement
ranges, lower resolutions, and work poorly outdoors due to
interference of the infrared spectrum of the sunlight with the
projected sensor pattern.

To the best of our knowledge, no publicly available dataset with laser scans at
the scale of the aforementioned vision benchmarks exists today.  Thus,
many recent Convolutional Neural Networks that are designed for Voxel
Grids \cite{brock2017gen,wu20153d} resort to artificially generated
data from the CAD models of ModelNet \cite{wu20153d}, a rather small,
synthetic dataset.  As a consequence, recent ensemble methods
(e.g.,~\cite{brock2017gen}) reach performance of over $97\%$ on
ModelNet10, which clearly indicates a model overfit due to limited
data.

%Some CNNs exploit the sparsity commonly found in voxel grids of laser scans, 
%for instance OctNet \cite{Riegler2017OctNet} or Vote3Deep \cite{Engelcke2017Vote3Deep}.
%The latter one is evaluated on the KITTI Vision Benchmark Suite \cite{geiger2012we}, 
%where it infers the 2D bounding box annotation of the images based on the provided laser scans. %in total 1.5 billion points, smaller than our data set

Those few existing laser scan datasets are mostly acquired with mobile
mapping devices or robots like \textit{DUT1}~\cite{zhuang2014novel},
\textit{DUT2}~\cite{zhuang2015contextual}, or
\textit{KAIST}~\cite{choe2013urban}, which are small ($<10^{7}$ points) and not publicly available.
%The Oakland data set \cite{munoz2009contextual} contains less than $2$ million labelled points. 
%
%Freiburg data set \cite{spinello2010layered}, \cite{spinello2011tracking} is around 10 million points.
%Indoor data sets \cite{lai2014unsupervised}, (\cite{anand2012contextually}, 
%\cite{koppula2011semantic}) contain only a few labelled rooms.
%
Publicly availabe laser scan datasets include the \textit{Oakland
  dataset}~\cite{munoz2009contextual} ($<2\times 10^{6}$ points), the
\textit{Sydney Urban Objects data set}~\cite{de2013unsupervised}, the \textit{Paris-rue-Madame database} \cite{serna2014paris} and
data from the \textit{IQmulus \& TerraMobilita
  Contest}~\cite{vallet2015terramobilita}. All have in common that
they use 3D LIDAR data from a mobile mapping car which provides a much
lower point density than a typical static scan, like ours.
They are also relatively small, such that supervised learning
algorithms easily overfit. The majority of today's available point
cloud datasets comes without a thorough, transparent evaluation that
is publicly available on the internet, continuously updated and that
lists all submissions to the benchmark.

With the \textit{semantic3D.net} benchmark presented in this paper, we
aim at closing this gap.  It provides the largest labeled 3D point
cloud data set with approximately four billion hand-labeled points,
comes with a sound evaluation, and continuously updates
submissions. It is the first dataset that allows full-fledged deep
learning on real 3D laser scans that have high-quality, manually
assigned labels per point.

\section{Objective}\label{sec:objective}
\vspace{-0.5em}

Given a set of points (here: dense scans from a static, terrestrial
laser scanner), we want to infer one individual class label per
point. We provide three baseline methods that are meant to represent
typical categories of approaches recently used for the task.

\emph{i)} \textit{2D image baseline:} 
%In many data sets it is possible to convert point clouds to (intensity) images or depth 
%images and apply standard techniques for semantic segmentation from image processing.

%\textcolor{red}{We need references that really work on depth images from laser scans, probably also 
%some that work on rgb-d. Only standard literature for image segmentation is not appropriate here....}

\begin{figure*}[th]
  \centering
  \includegraphics[width=\linewidth]{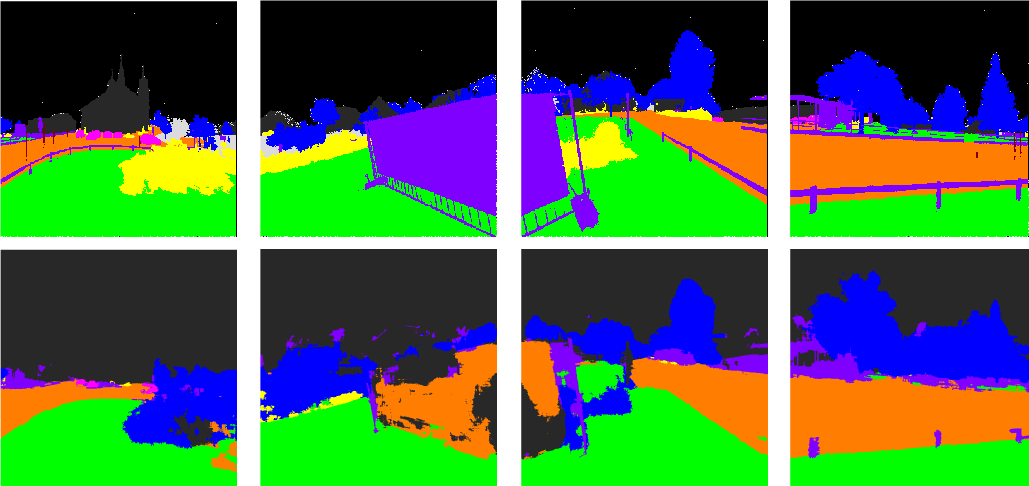}
   \caption{\emph{Top row:} projection of ground truth to
     images. \emph{Bottom row:} results of classification with the
     image baseline.  \emph{White:} unlabeled pixels, \emph{black:}
     pixels with no corresponding 3D point, \emph{gray:} buildings,
     \emph{orange:} man made ground, \emph{green:} natural ground,
     \emph{yellow:} low vegetation, \emph{blue:} high vegetation,
     \emph{purple: } hard scape, \emph{pink:} cars}
\label{fig:baseline1}
\end{figure*}

Many state-of-the-art laser scanners also acquire color values or even
entire color images for the scanned scene. Color images can add
additional object evidence that may help classification. Our first,
naive baseline classifies only the 2D color images without using any
depth information, so as to establish a link to the vast
literature on 2D semantic image segmentation.
%
%Early approaches make use of sliding windows to extract features, like HOG, for each pixel and use discriminative learning for inference \textcolor{red}{TODO: citations}.
%Properties like smoothness of class labels or frequencies of neighborhood patterns can be encoded, commonly by using Markov Random Fields. 
Modern methods use Deep Convolutional Neural Networks as a
workhorse. Encoder-decoder architectures, like SegNet
\cite{badrinarayanan2015segnet}, are able to infer the labels of an
entire image at once. Deep architectures can also be combined with
Conditional Random Fields (CRF) \cite{chen2016deeplab}.
%, and multiple tasks can be addressed at once, like edge detection \cite{zimmerman1998ubernet,chen2015semantic}.
Our baseline method in Section \ref{subsec:image_baseline} covers image-based semantic segmentation.

\emph{ii)} \textit{3D Covariance baseline:} A more specific approach,
which takes advantage of the 3D information, is to work on point
clouds directly.  We use a recent implementation of the standard
classification pipeline, i.e., extract hand-crafted features from 3D
(multi-scale) neighbourhoods, and feed them to a discriminative
learning algorithm. Typical features encode surface properties based
on the covariance tensor of a point's neighborhood \cite{demantke2011}
or a randomized set of histograms \cite{blomley2014}. Additionally,
height distributions can be encoded by using cylindrical neighborhoods
\cite{monnier2012,weinmann2013feature}. The second baseline method
(Section \ref{subsec:covariance_baseline}) represents this category.

\emph{iii)} \textit{3D CNN baseline:} It is a rather obvious extension
to apply deep learning also to 3D point clouds, mostly using voxel
grids to obtain a regular neighbourhood structure. To work efficiently
with large point neighborhoods in clouds with strongly varying
density, recent work uses adaptive neighbourhood data structures like
octrees~\cite{wu20153d,brock2017gen,Riegler2017OctNet} or sparse voxel
grids~\cite{Engelcke2017Vote3Deep}. Our third baseline method in
Section \ref{subsec:deep_net_baseline} is a basic, straight-forward
implementation of 3D voxel-grid CNNs.

\subsection{2D Image Baseline}\label{subsec:image_baseline}
\vspace{-0.5em}

We convert color values of the scans to separate images (without
depth) with cube mapping \cite{greene1986environment}. Ground truth
labels are also projected from the point clouds to image space, such
that the 3D point labeling task turns into a pure semantic image
segmentation problem of 2D RGB images (Figure~\ref{fig:baseline1}). We chose the associate
hierarchical fields method~\cite{ladicky13} for semantic segmentation
because it has proven to deliver good performance for a variety of
tasks (e.g.,~\cite{montoya2014,ladicky2014eccv}) and was available in
its original implementation.

The method works as follows: four different types of features --
texton~\cite{malik01}, SIFT~\cite{lowe04}, local quantized ternary
patters~\cite{hussain12} and self-similarity
features~\cite{shechtman07} -- are extracted densely per image
pixel. Each feature category is separately clustered into $512$
distinct patterns using standard K-means clustering, which
corresponds to a typical bag-of-words representation.
For each pixel in an image, the feature vector is a concatenation of
bag-of-word histograms over a fixed set of $200$ rectangles of varying
sizes. These rectangles are randomly placed in an extended
neighbourhood around a pixel.
We use multi-class boosting~\cite{torralba04} as classifier and the
most discriminative weak features are found as explained
in~\cite{shotton06}.
To add local smoothing without loosing sharp object boundaries, we
smooth inside superpixels and favor class transitions at their
boundaries.
Super-pixels are extracted via mean-shift~\cite{comaniciu02} with 3
sets of coarse-to-fine parameters as described
in~\cite{ladicky13}. Class likelihoods of overlapping superpixels are
predicted using the feature vector consisting of a bag-of-words
representation of each superpixel. Pixel-based and superpixel-based
classifiers with additional smoothness priors over pixels and
superpixels are combined in a probabilistic fashion in a conditional
random field framework as proposed in~\cite{kohli08}. The most
probable solution of the associative hierarchical optimization problem
is found using the move making~\cite{boykov01} graph-cut based
algorithm~\cite{boykov04}, with appropriate graph construction for
higher-order potentials~\cite{ladicky13}.

%Finally, the classification results in image space are mapped to the point cloud by again deploying cube mapping.
%
%Please note that scans and color values were not recorded at exactly the same time, which causes mismatches for moving objects. This becomes particularly apparent for classes \emph{scanning artefacts}\footnote{often caused by pedestrians moving through the scene while scan acquisition} and \emph{cars}.

\subsection{3D Covariance Baseline}\label{subsec:covariance_baseline}
\vspace{-0.5em}
%summary of baseline 2, more details can be read in the paper
%state of the art for this approach

The second baseline was inspired by \cite{weinmann2015distinctive}.
It infers the class label directly from the 3D point cloud using
multiscale features and discriminative learning. Again, we had access
to the original implementation.
That method uses an efficient approximation of multi-scale
neighbourhoods, where the point cloud is sub-sampled into a
multi-resolution pyramid, such that a constant, small number of
neighbours per level captures the multi-scale information.
The multi-scale pyramid is generated by voxel-grid filtering with
uniform spacing.

The feautre set extracted at each level is an extension of the one
decribed in~\cite{weinmann2013feature}. It uses different combinations
of eigenvalues and eigenvectors of the covariance per
point-neighborhood to different geometric surface
properties. Furthermore, height features based on vertical,
cylindrical neighbourhoods are added to emphasize the special role of
the gravity direction (assuming that scans are, as usual, aligned to
the vertical).
Note that we do not make use of color values or scanner
intensities. These are not always available in point clouds, and we
empirically found that they do not improve the results of the
method. As classifier, we use a random forest, where optimal
parameters are found with grid search and five fold cross-validation.
Please refer to \cite{hackel2016b} for details.

\subsection{3D CNN Baseline}\label{subsec:deep_net_baseline}
\vspace{-0.5em}

\begin{figure}[th]
  \centering
    \vspace{0.3cm}
    \subfloat{%[\label{fig:data_intensity}]{%
       \includegraphics[width=0.49\textwidth]{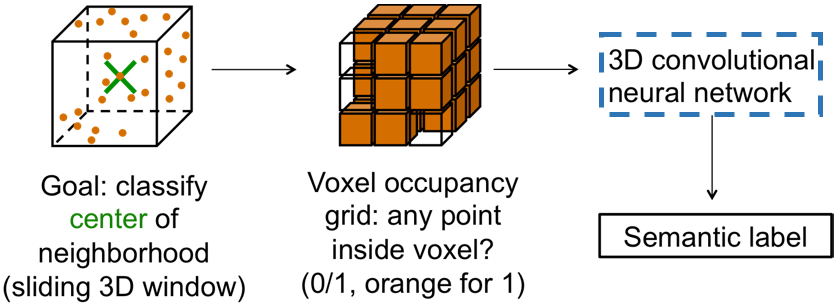}
    }
    \vspace{0.7cm}
    %\hfill
    \subfloat{%[\label{fig:data_rgb}]{%
       \includegraphics[width=0.49\textwidth]{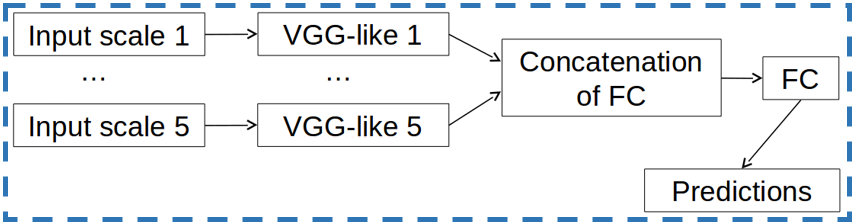}
    }
   \caption{Our deep neural network pipeline.}
\label{fig:pipeline}
\end{figure}

We design our baseline for the point cloud classification task
following recent VoxNet \cite{maturana2015voxnet} and ShapeNet
\cite{wu20153d} 3D encoding ideas. The pipeline is illustrated in
Fig.~\ref{fig:pipeline}.
Instead of generating a global 3D voxel-grid prior to processing, we
create $16\times16\times16$ voxel cubes per scan
point\footnote{This strategy automatically centers each voxel-cube per scan point. Note 
that for the alternative approach of a global voxel grid, several scan points 
could fall into the same grid cell in dense scan parts. This would require scan point 
selection per grid cell, which is computationally costly and results in (undesired) down-sampling.}. 
%
%In order to assign a label to each point individually we follow a sliding windows (or rather cube) approach, 
%where a dense $16\times16\times16$ voxel grid is slid over the point cloud. 
%
%This dense 3D grid represents the input data of our deep 3D convolutional network (CNN). In order to represent very large as well as very small objects, multiple scales are used for the generation of the voxel grids.
We do this at $5$ different resolutions, where voxel size ranges from
$2.5$ cm to $40$ cm (multiplied by powers of $2$) and encode empty voxel cells as $0$ and filled ones as $1$. 
The input to the CNN is thus encoded in a multidimensional tensor with $5 \times16\times16\times16$ cube entries per scan point.
\vspace{-0.5em}

Each of the five scales is handled separately by a VGG-like network
path which includes convolutional, pooling and ReLU layers.
% are fed with the $16\times16\times16$ representations.
%
The $5$ separate network paths are finally concatenated into a
single representation, which is passed through two fully-connected
layers.
The output of the second fully-connected layer is an $8$-dimensional
vector, which contains the class scores for each of the $8$ classes in
this benchmark challenge. Scores are transformed to class conditional
probabilities with the soft-max function.
\vspace{-0.5em}

Before describing the network architecture in detail we introduce the
following notation:
\vspace{-0.5em}

%\begin{itemize}
$c(i, o)$ stands for convolutional layers with $3\times 3 \times
  3$ filters, $i$ input channels, $o$ output channels, zero-padding of
  size $1$ at each border and a stride of size $1$.  %The number of parameters of this convolutional layer is: $o / i \times (27 + 1)$, where the $1$ is for the biases,
$f(i, o)$ stands for fully-connected layers.
$r$ stands for a ReLU non-linearity,
$m$ stands for a volumetric max pooling with receptive field $2 \times 2 \times 2$, applied with a stride of size $2$ in each dimension,
$d$ stands for a dropout with $0.5$ probability,
$s$ stands for a soft-max layer.
%\end{itemize}

Our 3D CNN architecture assembles these components to a VGG-like
network structure.
%powerful learning engine.% deep net is a tuple of such entries. 
We choose the filter size in convolutional layers as small as possible
($3 \times 3 \times 3$), as recommended in recent work \cite{he2016deep}, to have the
least amount of parameters per layer and, hence, reduce both the risk of overfitting
and the computational cost.
\vspace{-0.5em}

For the $5$ separate network paths that act on different resolutions, we use a VGG-like~\cite{simonyan2014very} architecture: 
\begin{equation*}
(c(1, 16), r, m, c(16, 32), r, m, c(32, 64), r, m). 
\end{equation*}
The output is vectorized, 
concatenated between scales and the two fully-connected layers are applied on top to predict the class responses:
\begin{equation*}
(f(2560, 2048), r, d, f(2048, 8), s).
\end{equation*}
For training we deploy the standard multi-class cross-entropy loss.
%
%Its main benefit is that erroneaus predictions with a high
%class-conditional probability incur a higher penalty than errors
%where the classifier is uncertain and only assigns small
%probabilities.
%
%During test time we compute the classification loss; a small classification loss is the objective of this benchmark.
%
Deep learning is non-convex but it can be efficiently optimized with stochastic gradient descent
(SGD), which produces classifiers with state-of-the-art prediction performance.
The SGD algorithm uses randomly sampled mini-batches of several
hundred points per batch to iteratively update the parameters of the
CNN. We use the popular adadelta algorithm~\cite{zeiler2012adadelta} for optimization, an extension of
stochastic gradient decent~\cite{bottou2010large}. 
%One major advantage of adadelta is that we do not need to set a learning rate and it turns out to be robust to large gradients and noise.

We use a mini batch size of $100$ training samples (i.e., points),
where each batch is sampled randomly and balanced (contains equal
numbers of samples per class).  We run training for $74{,}700$ batches
and sample training data from a large and representative point cloud
with $259$ million points (sg28\_4).
A standard pre-processing step for CNNs is data augmentation to
enlarge the training set and to avoid overfitting.
Here, we augment the training set with a random rotation around the
z-axis after every $100$ batches.
% because the voxel data representation is not invariant to rotations..
During experiments it turned out that additional training data did not
improve performance. This indicates that in our case we rather deal
with underfitting (as opposed to overfitting), i.e. our model lacks
the capacity to fully capture all the evidence in the available training
data\footnote{Note that our model reaches hardware limits of our GPU
  (TitanX with 12GB of RAM) and we thus did not experiment with larger
  networks at this point.}. We thus refrain from further possible
augmentations like randomly missing points or adding noise.
%However, it turned out that these additional augmentations did not improve  
%
%Often, the training process is stopped to avoid overfitting. 
%We did not perform early stopping and continued training until the error converged. 
\vspace{-0.5em}

Our network is implemented in C++ and Lua and uses the Torch7
framework~\cite{torch} for deep learning. The code and the documentation for this baseline are
publicly available at \\
\url{https://github.com/nsavinov/semantic3dnet}.

\begin{figure*}[th]
\centering
    \subfloat{%[\label{fig:data_intensity}]{%
       \includegraphics[width=0.32\textwidth]{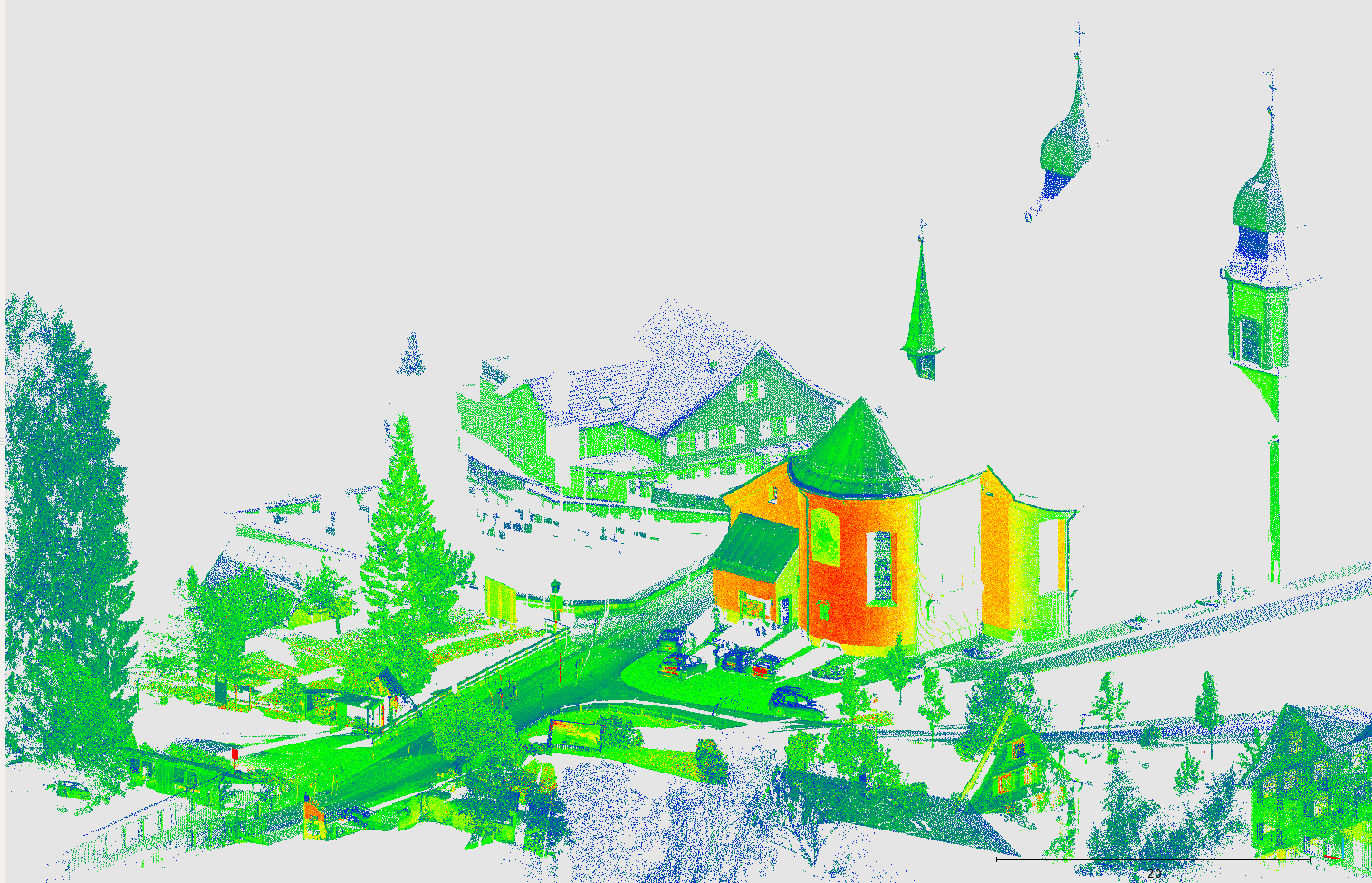}
    }
    \hfill
    \subfloat{%[\label{fig:data_rgb}]{%
       \includegraphics[width=0.32\textwidth]{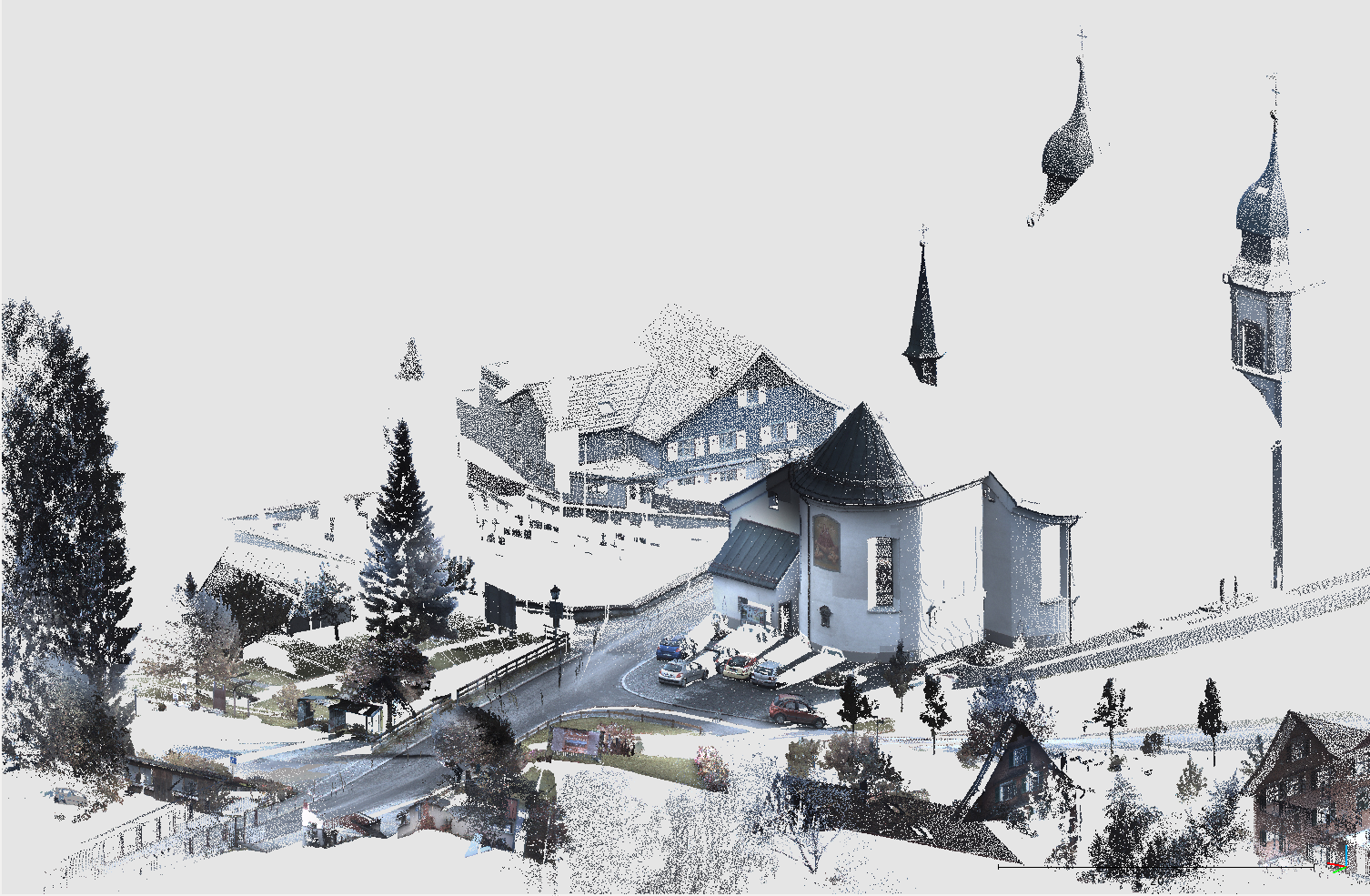}
    }
    \hfill
    \subfloat{%[\label{fig:data_classes}]{%
       \includegraphics[width=0.32\textwidth]{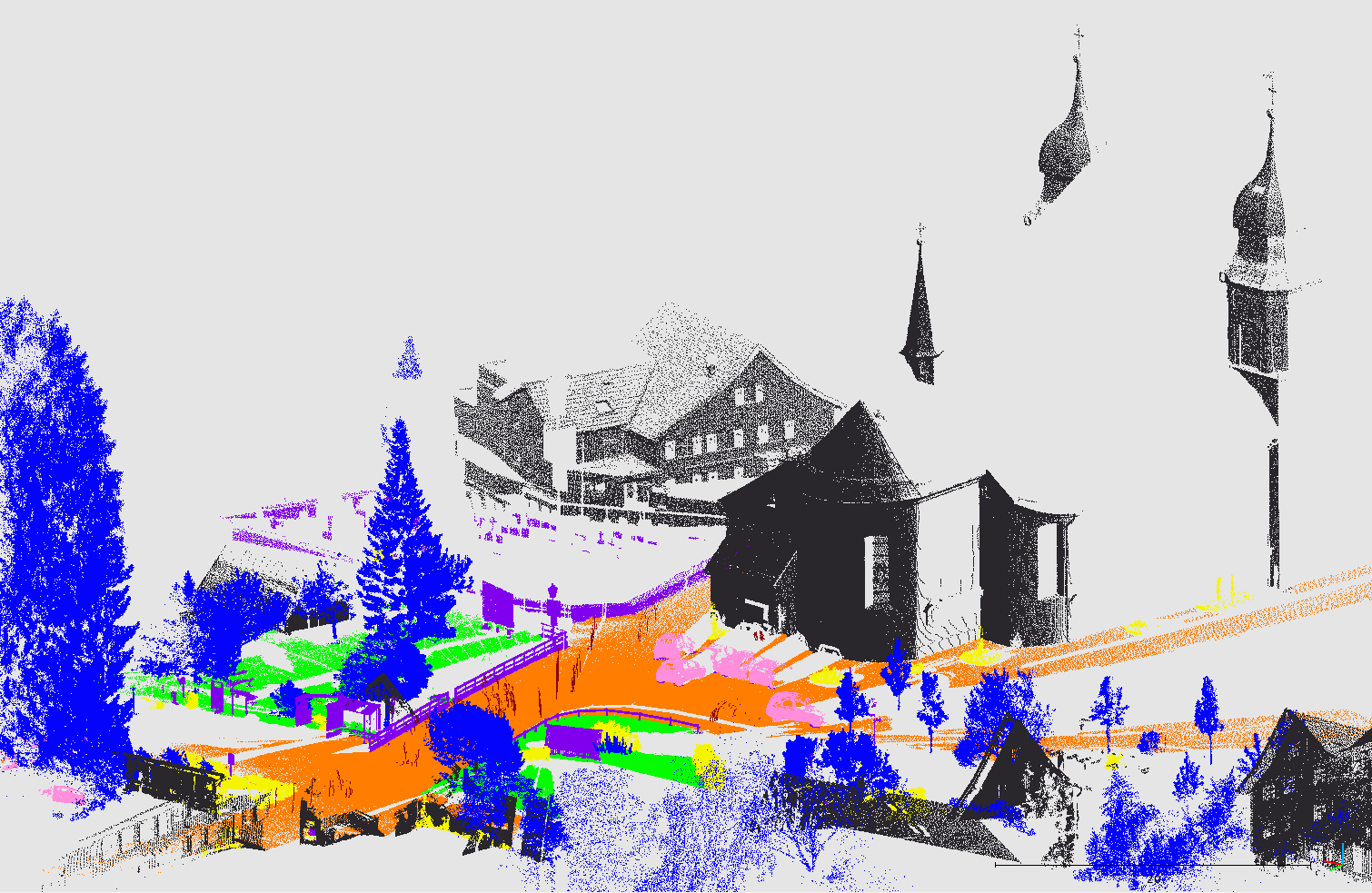}
    }
    \vfill
    \subfloat{%[\label{fig:data_intensity}]{%
       \includegraphics[width=0.32\textwidth]{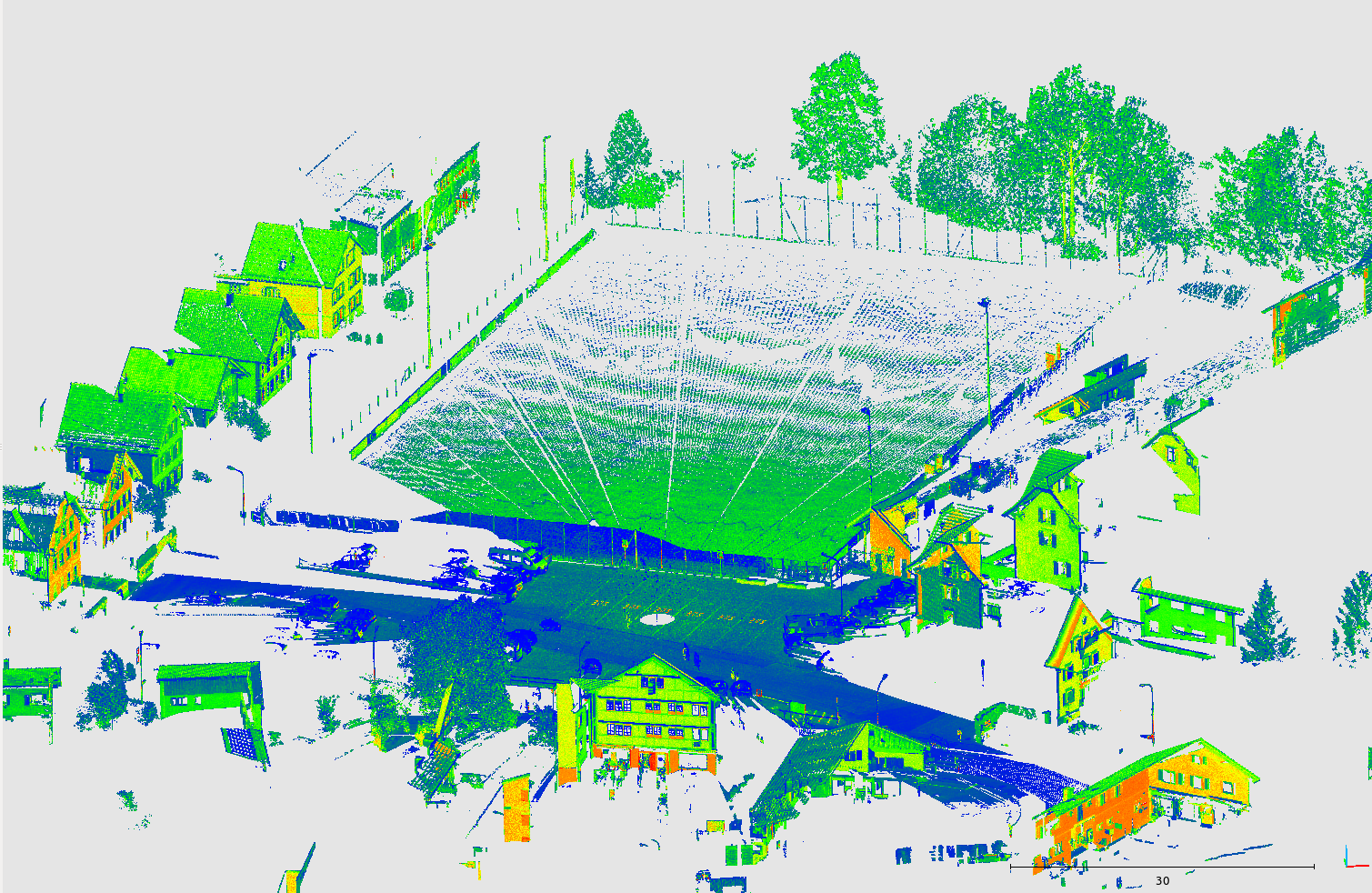}
    }
    \hfill
    \subfloat{%[\label{fig:data_rgb}]{%
       \includegraphics[width=0.32\textwidth]{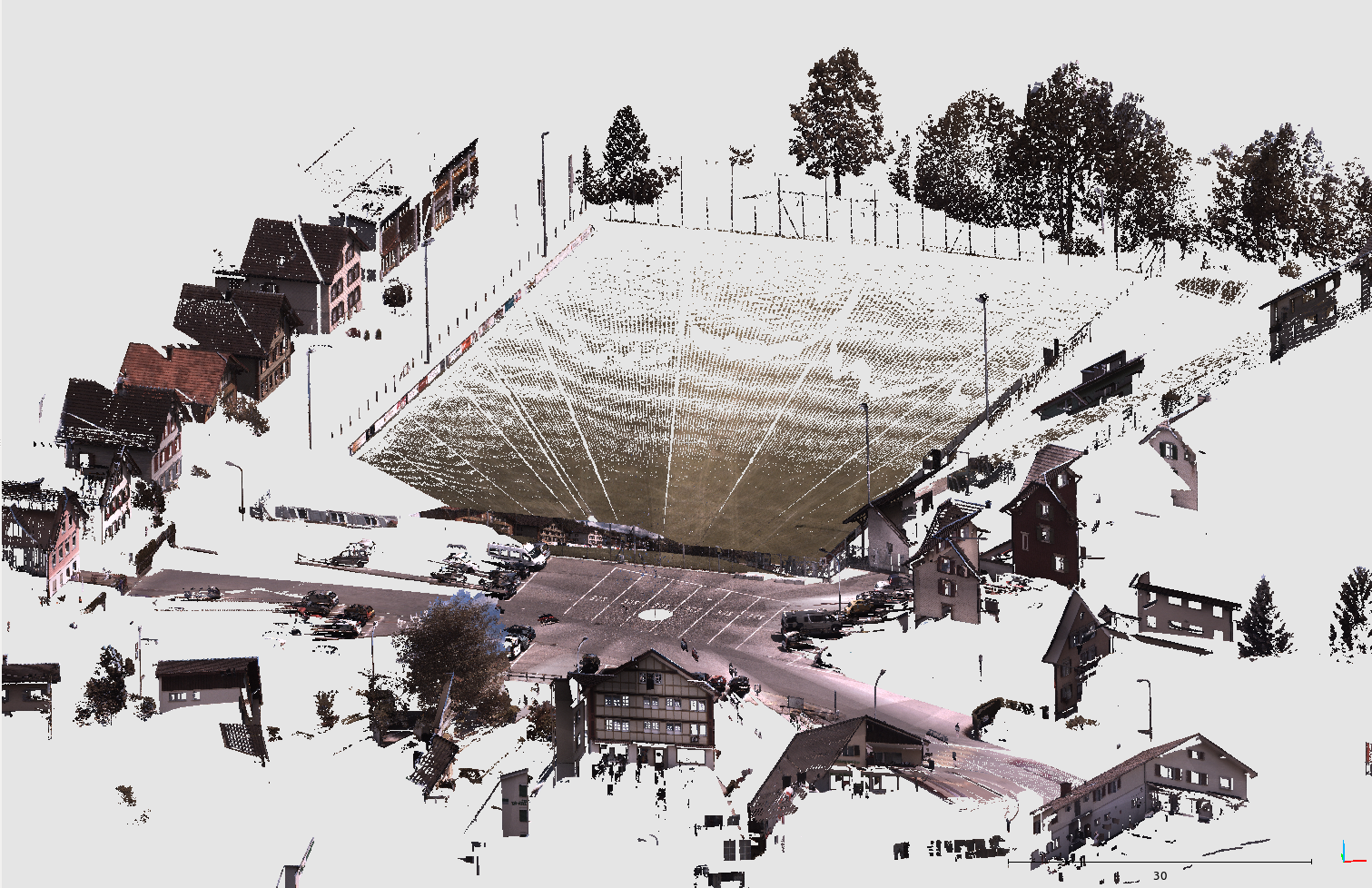}
    }
    \hfill
    \subfloat{%[\label{fig:data_classes}]{%
       \includegraphics[width=0.32\textwidth]{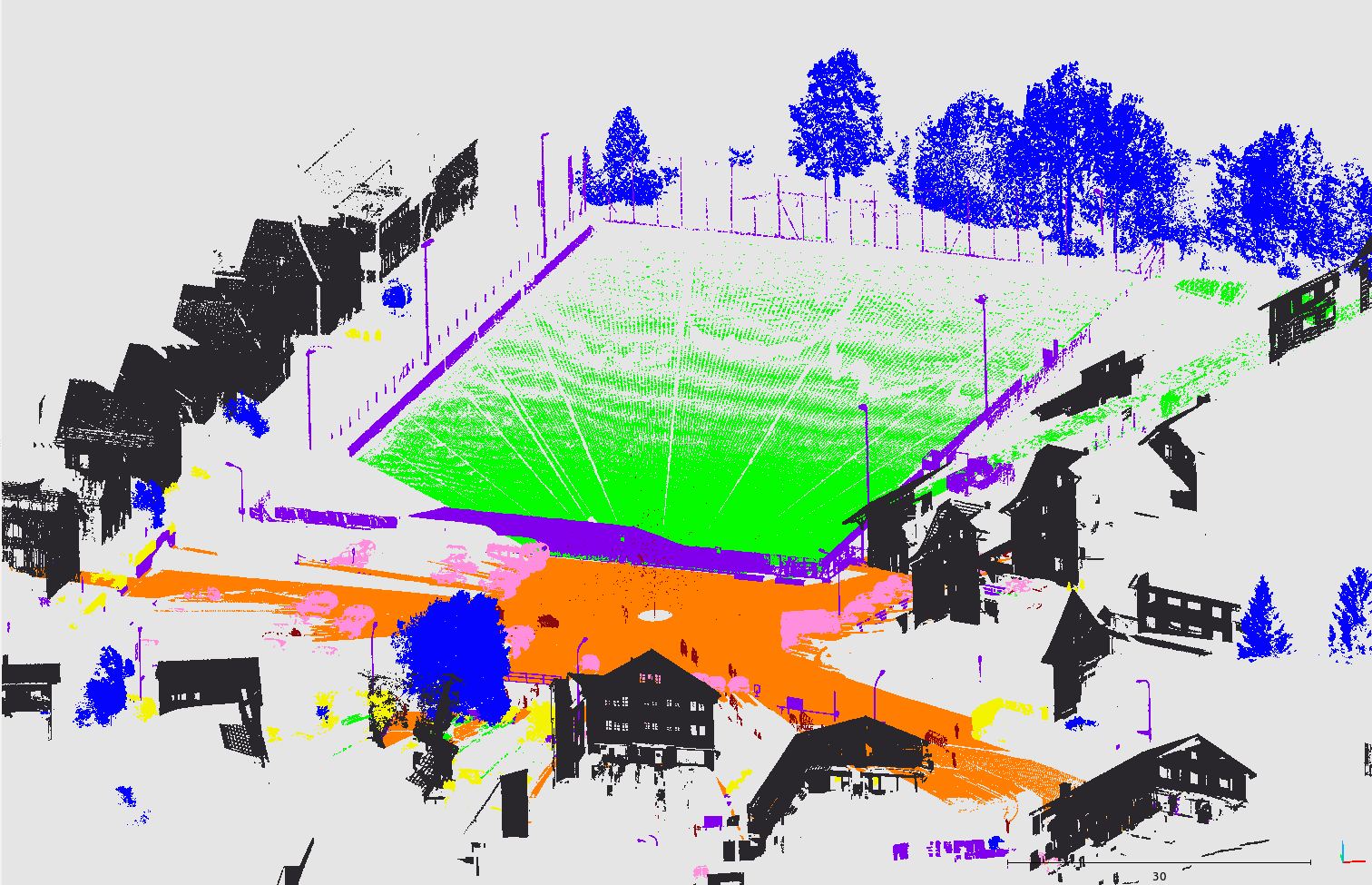}
    }
    \vfill
    \subfloat{%[\label{fig:data_intensity}]{%
       \includegraphics[width=0.32\textwidth]{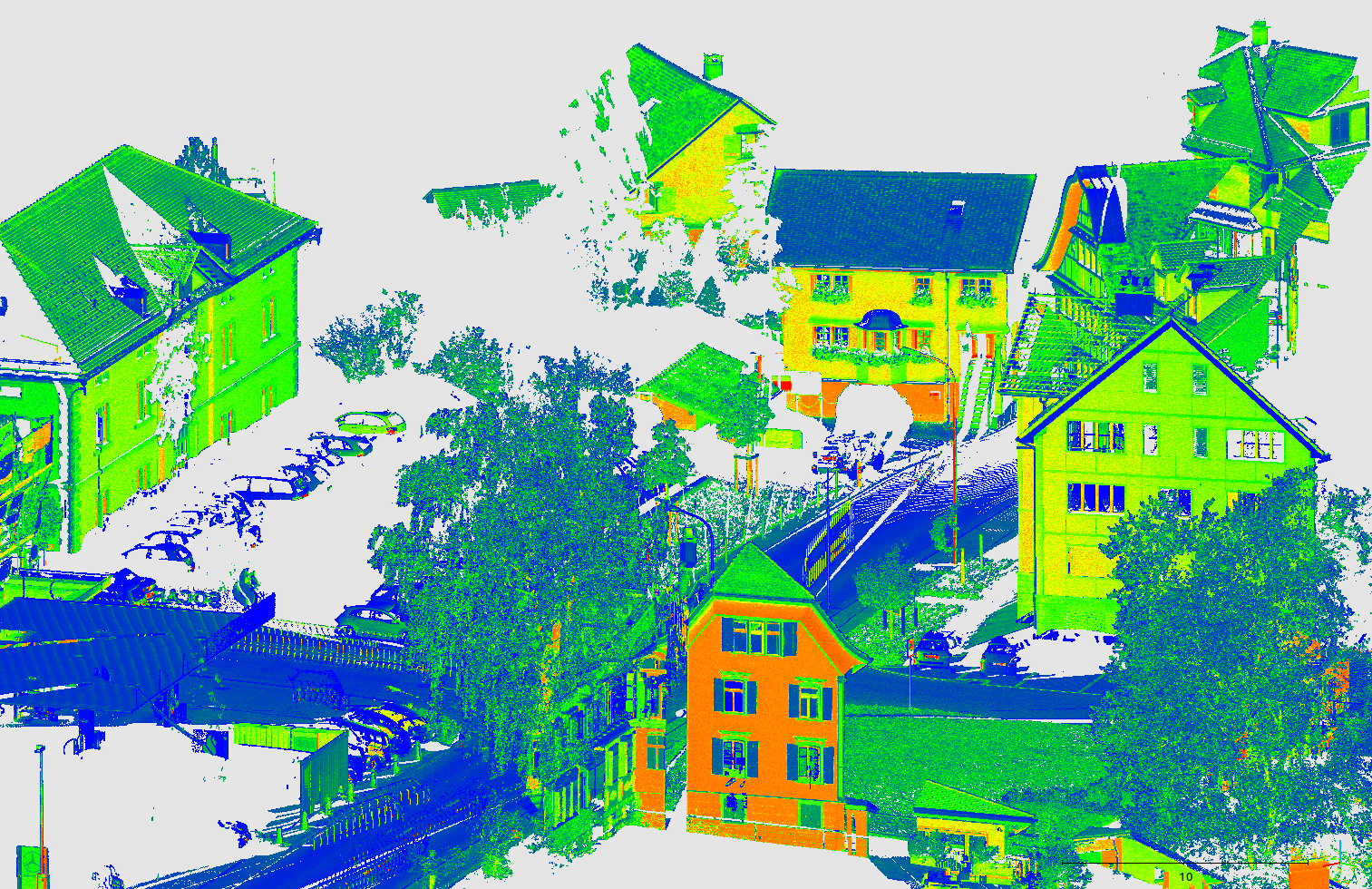}
    }
    \hfill
    \subfloat{%[\label{fig:data_rgb}]{%
       \includegraphics[width=0.32\textwidth]{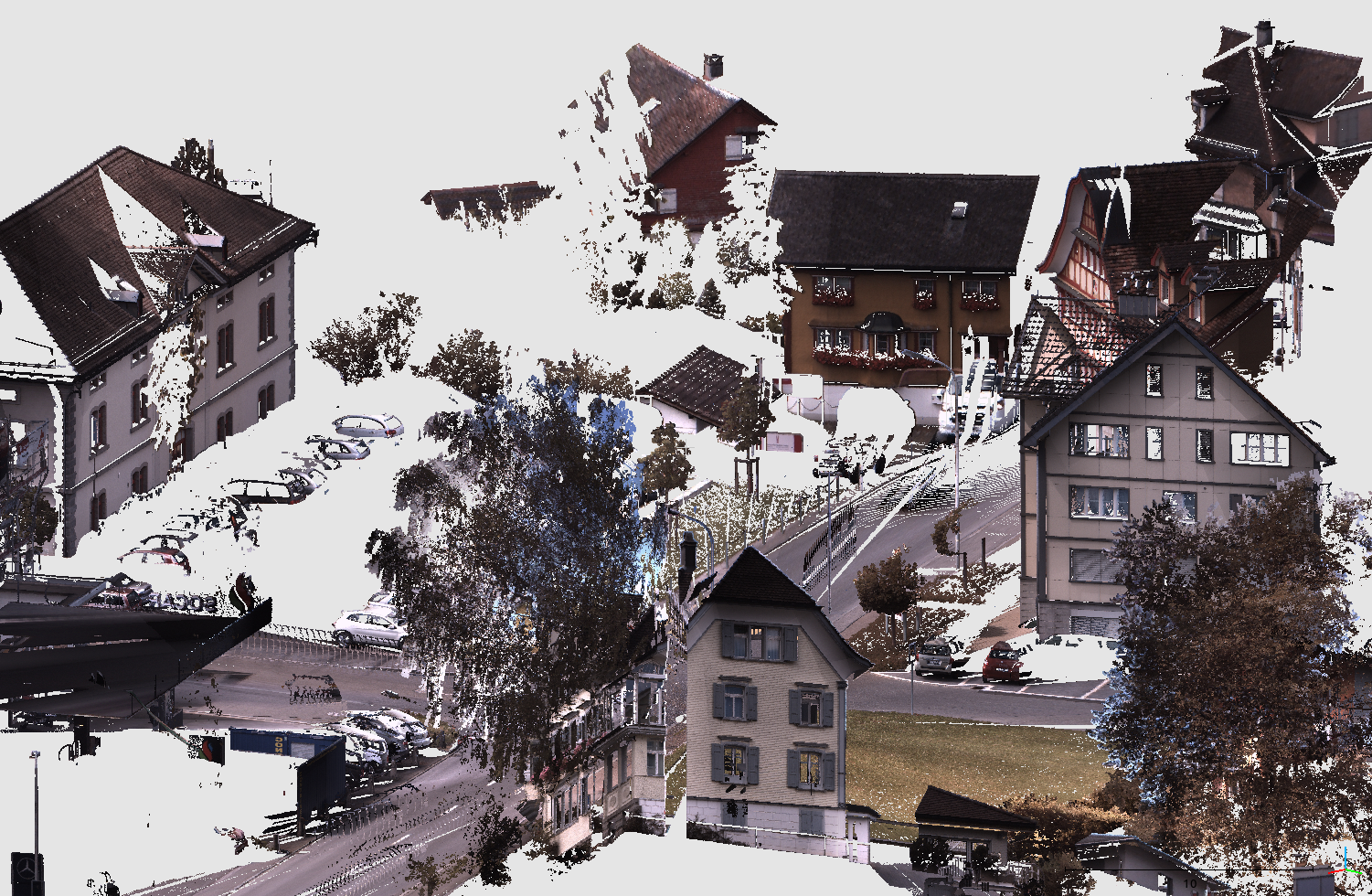}
    }
    \hfill
    \subfloat{%[\label{fig:data_classes}]{%
       \includegraphics[width=0.32\textwidth]{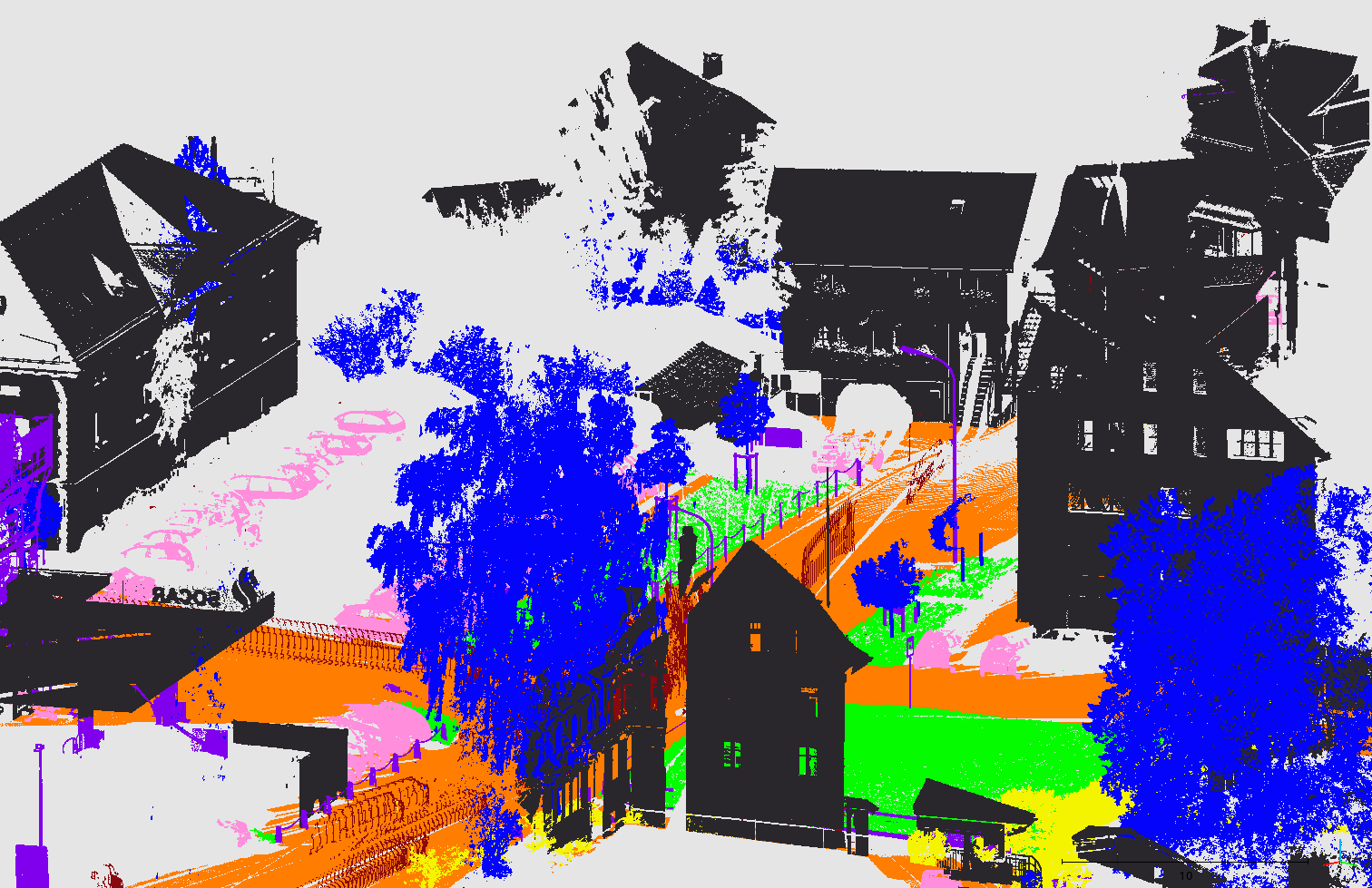}
    }
    \caption{Intensity values (left), rgb colors (middle) and class labels (right) for example data sets.}
      \label{fig:data_channels}
\end{figure*}

\section{Data}\label{sec:data}
\vspace{-0.5em}

%Beside future technologies like autonomous cars there already exists a large market for laser scanners,
%which has evolved into three fields with distinct product specifications: 
%airborne scanning, mobile mapping vehicles and terrestrial laser scanning.
 
%For the task of object detection terrestrial laser scans are possibly the most challenging data sets,
%not only due to the broad range of possible applications, from city modelling over monitoring to mining.
%They have a very high angular resolution and larger measurement ranges than mobile mapping vehicles, 
%so that it is possible to detect not only large objects, like buildings, but also very tiny structures.

Our $30$ published terrestrial laser scans consist of in total
$\approx 4$~billion 3D points and contain urban and rural scenes, like
farms, town halls, sport fields, a castle and market squares. We
intentionally selected various different natural and man-made scenes
to prevent over-fitting of the classifiers.  All of the published
scenes were captured in Central Europe and describe typical European
architecture as shown in Figure
\ref{fig:data_channels}. Surveying-grade laser scanners were used for
recording these scenes. Colorization was performed in a post
processing step by deploying a high resolution cubemap, which was
generated from camera images. In general, static laser scans have a
very high resolution and are able to measure long distances with
little noise. Especially compared to point clouds derived via
structure-from-motion pipelines or Kinect-like structured light
sensors, laser scanners deliver superior data quality.
\vspace{-0.5em}

Scanner positions for data recording were selected as usually done in
the field: only little scan overlap as needed for registration, so
that scenes can be recorded in a minimum of time. This free choice of the
scanning position implies that no prior assumption based on point
density and on class distributions can be made.
%what makes training more challenging than data 
%from mobile mapping systems, where for instance houses typically are close to streets. 
%As a side-effect, our scenes are not only suited for evaluating semantic classifiers but could 
%also be used as a challenging test set for registration algorithms.
%
We publish up to $3$ laser scans per scene that have small
overlap. The relative position of laser scans at the same location was
estimated from targets.
\vspace{-0.5em}

We use the following $8$ classes within this benchmark challenge, which cover:
\emph{1)} man made terrain: mostly pavement; 
\emph{2)} natural terrain: mostly grass;
\emph{3)} high vegetation: trees and large bushes; 
\emph{4)} low vegetation: flowers or small bushes which are smaller than $2$ m; 
\emph{5)} buildings: Churches, city halls, stations, tenements, etc.;
\emph{6)} remaining hard scape: a clutter class with for instance garden walls, 
fountains, banks, etc.;
\emph{7)} scanning artifacts: artifacts caused by dynamically moving objects during the 
recording of the static scan;
\emph{8)} cars and trucks.
Some of these classes are ill-defined, for instance some scanning artifacts could also go 
for cars or trucks and it can be hard to differentiate between large and small bushes. 
Yet, these classes can be helpful in numerous applications. Please note that in most 
applications class \emph{7}, scanning artifacts, gets filtered with heuristic rule sets. Within this 
benchmark we want to deploy machine learning techniques instead, and thus do not perform any heuristic pre-processing.
\vspace{-0.5em}

In our view, large data sets are important for two reasons: \emph{a)}
Typically, real world scan data sets are large. Hence, methods which
have an impact on real problems have to be able to process a large
amount of data.  \emph{b)} Large data sets are especially important
when developing methods with modern inference techniques capable of
representation learning.  With too small datasets, good results leave
strong doubts about possible overfitting; unsatisfactory results, on
the other hand, are hard to interpret as guidelines for further
research: are the mistakes due to short-comings of the method, or
simply caused by unsufficient training data?

\subsection{Point Cloud Annotation}
\vspace{-0.5em}
In contrast to common strategies for 3D data labelling that first
compute an over-segmentation followed by segment-labeling, we manually
assign each point a class label individually.  Although this strategy
is more labor-intensive, it avoids inheriting errors from the
segmentation approach and, more importantly, classifiers do not learn
hand-crafted rules of segmentation algorithms when trained with the
data.
In general, it is more difficult to label a point cloud by hand than
images. The main problem is that it is hard to select a 3D point on a
2D monitor from a set of millions of points without a clear
neighbourhood/surface structure.
We tested two different strategies:
\vspace{-0.5em}
\paragraph{Annotation in 3D:}
We follow an iterative filtering strategy, where we manually select a
couple of points, fit a simple model to the data, remove the model
outliers and repeat these steps until all inliers belong to the same
class. With this procedure it is possible to select large buildings in
a couple of seconds. A small part of the point clouds was labeled with 
this approach by student assistants at ETH Zurich. 
\vspace{-0.5em}
%Although this semi-automated human-in-the-loop
%classification strategy is very efficient, no publicly available
%software exists that could easily be shared with external workers for
%labeling points. \textcolor{red}{I don't get that? Why couldn't we
%  write the software ourselves and then share it?}\textcolor{blue}{The point clouds at ETH were labeled with the 3D tool. For Sascha it appeared to be more convenient to label 
%with leica cyclone.}

\paragraph{Annotation in 2D:}
The user rotates a point cloud, fixes a 2D view and draws a closed
polygon which splits a point cloud into two parts (inside and outside
of the polygon). One part usually contains points from the background
and is discarded. This procedure is repeated a few times until all
remaining points belong to the same class. At the end all points are
separated into different layers corresponding to classes of
interest. This 2D procedure works well with existing software packages
\cite{CloudCompare} such that it can be outsourced to external
labelers more easily than the 3D work-flow. We used this procedure 
for all data sets where annotation was outsourced.

\begin{table*}[thp]
  \vspace{+0.5em}
  \centering
  \resizebox{\textwidth}{!}{%
  \begin{tabular}{l|ccc|cccccccc}
   { Method} & $\overline{ IoU }$ & $OA$ & $t [s]$ & $IoU_1$ & $IoU_2$ & $IoU_3$ & $IoU_4$ & $IoU_5$ & $IoU_6$ & $IoU_7$ & $IoU_8$\\
   \hline

   HarrisNet  & {\bf0.623} & 0.881      & unknown & 0.818     & 0.737     & {\bf0.742}& {\bf0.625}& {\bf0.927}& 0.283     & 0.178     & {\bf0.671} \\
   DeepSegNet & 0.516      & {\bf0.884} & unknown & 0.894     & {\bf0.811}& 0.590     & 0.441     & 0.853     & {\bf0.303}& {\bf0.190}& 0.050      \\
   TMLC-MS    & 0.494      & 0.850      &  38421  & {\bf0.911}& 0.695     & 0.328     & 0.216     & 0.876     & 0.259     & 0.113     & 0.553      \\
   TML-PC     & 0.391      & 0.745      & unknown & 0.804     & 0.661     & 0.423     & 0.412     & 0.647     & 0.124     & 0.0*        & 0.058
  \end{tabular}}
  \caption{Semantic3d benchmark results on the full data set: 3D covariance baseline \textit{TMLC-MS}, 2D RGB image baseline \textit{TML-PC}, and first submissions \textit{HarrisNet} and \textit{DeepSegNet}. \textit{IoU} for categories (1) man-made terrain, (2) natural terrain, (3) high vegetation, (4) low vegetation, (5) buildings, (6) hard scape, (7) scanning artefacts, (8) cars. * Scanning artefacts were ignored for 2D classification because they are not present in the image data.}%.}
  \label{tab:sem3d_full}
  \vspace{0.5em} % improve text placement on last page
\end{table*}
\begin{table*}[thp]
  \vspace{+0.5em}
  \centering
  \resizebox{\textwidth}{!}{%
  \begin{tabular}{l|ccc|cccccccc}
   { Method} & $\overline{ IoU }$ & $OA$ & $t [s]$ & $IoU_1$ & $IoU_2$ & $IoU_3$ & $IoU_4$ & $IoU_5$ & $IoU_6$ & $IoU_7$ & $IoU_8$\\
   \hline
   TMLC-MSR & {\bf0.542} & {\bf0.862}& 1800   & {\bf0.898}& {\bf0.745}& 0.537     & {\bf0.268}& {\bf0.888}& {\bf0.189}& {\bf0.364}& {\bf0.447} \\
   DeepNet  & 0.437      & 0.772     & 64800  & 0.838     & 0.385     & {\bf0.548}& 0.085     & 0.841     & 0.151     & 0.223     & 0.423 \\
   TML-PCR  & 0.384      & 0.740     & unknown& 0.726     & 0.73      & 0.485     & 0.224     & 0.707     & 0.050     & 0.0*       & 0.15
  \end{tabular}}
  \caption{Semantic3d benchmark results on the reduced data set: 3D covariance baseline \textit{TMLC-MSR}, 2D RGB image baseline \textit{TML-PCR}, and our 3D CNN baseline \textit{DeepNet}. \textit{TMLC-MSR} is the same method as \textit{TMLC-MS}, the same goes for \textit{TMLC-PCR} and \textit{TMLC-PC}. In both cases \textit{R} indicates classifiers on the reduced dataset. \textit{IoU} for categories (1) man-made terrain, (2) natural terrain, (3) high vegetation, (4) low vegetation, (5) buildings, (6) hard scape, (7) scanning artefacts, (8) cars. * Scanning artefacts were ignored for 2D classification because they are not present in the image data.}
  \label{tab:sem3d_reduced}
  \vspace{0.5em} % improve text placement on last page
\end{table*}

\section{Evaluation}
\vspace{-0.5em}
\label{sec:evaluation}

We follow Pascal VOC challenge's \cite{everingham2010pascal} choice of
the main segmentation evaluation measure and use \textit{Intersection
  over Union} ($IoU$)\footnote{$IoU$ compensates for different class frequencies as opposed to, for example, 
\emph{overall accuracy} that does not balance different class frequencies giving higher influence to large classes.} averaged over all classes. 
%$F_1$ Scores have a larger weight on true positives than on errors and.
%Hence, $F_1$ Scores are less strict on the evaluation of small classes, 
%where usually the contribution of type II errors is large (Please check: weight should be two times larger?). 
%
Assume classes are indexed with integers from $\{1, \dots, N\}$ where
$N$ is an overall number of classes.  Let $C$ be an $N \times N$
confusion matrix of the chosen classification method, where each entry
$c_{ij}$ is a number of samples from ground-truth class $i$ predicted
as class $j$. Then the evaluation measure per class $i$ is defined as
\begin{equation}
    IoU_i = \frac{c_{ii}}{c_{ii} + \sum\limits_{j \neq i} c_{ij} + \sum\limits_{k \neq i} 
c_{ki}}.
\end{equation}
The main evaluation measure of our benchmark is thus
\begin{equation}
    \overline{IoU} = \frac{\sum\limits_{i = 1}^N IoU_i}{N}.
\end{equation}

We also report $IoU_i$ for each class $i$ and overall accuracy
\begin{equation}
    OA = \frac{\sum\limits_{i = 1}^N c_{ii}}{\sum\limits_{j = 1}^N \sum\limits_{k = 1}^N 
c_{jk}}
\end{equation}
as auxiliary measures and provide the confusion matrix $C$.
Finally, each participant is asked to specify the time $T$ it took to
classify the test set as well as the hardware used for
experiments. This measure is important for understanding how suitable
the method is in real-world scenarios where usually billions of points
are required to be processed.

For computational demanding methods we provide a reduced challenge
consisting of a subset of the published test data.  The results of our
baseline methods as well as submissions are shown in Table
\ref{tab:sem3d_full} for the full challenge and in Table
\ref{tab:sem3d_reduced} for the reduced challenge.
Of the three published baseline methods the covariance based method
performs better than the CNN baseline and the color based method.  Due
to its computational cost we could only run our own deep learning
baseline \textit{DeepNet} on the reduced data set. We expect a network
with higher capacity to perform much better. Results on the full
challenge of two (unfortunately yet unpublished) 3D CNN methods,
\textit{DeepSegNet} and \textit{HarrisNet}, already beat our
covariance baseline by a significant margin
(Table~\ref{tab:sem3d_full}) of $2$ respective $12$ percent points.
%
%\textcolor{red}{Discuss a bit more! e claim earlier that we
%  ``discuss'' first submissions. But we really only mention them. Any
%  intuition what exactly they do? Why it works well? In any case, at
%  least comment that as expected, deep learning seems to be the way to
%  go also for point clouds, if you have enough data. So our benchmark
%  already starts to work and generate progress}
%The two deep learning submissions show that when sufficiently large labeled point clouds are provided, deep learning seems to be the way to go.
This indicates that deep learning seems to be the way to go also for point clouds, 
if enough data is available for training. It is a first sign that our 
benchmark already starts to work and generates progress.

\begin{figure*}[th!]
\centering
    \subfloat[\label{fig:class_distr}]{%
       \includegraphics[width=0.47\textwidth]{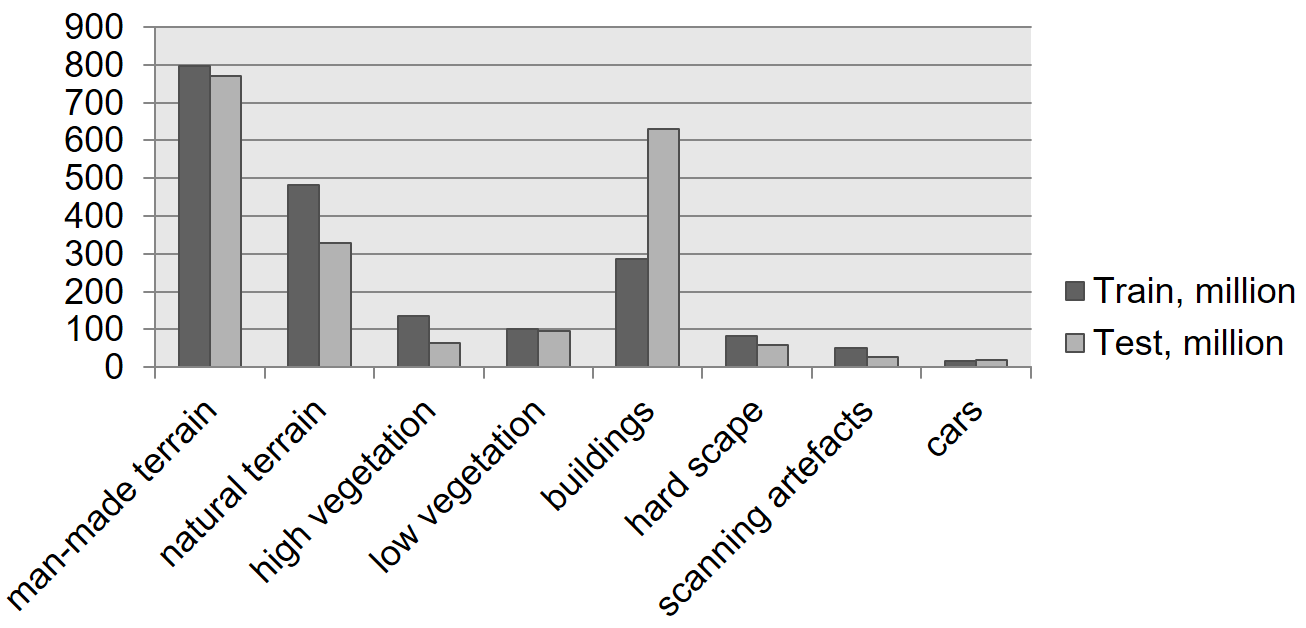}
    }
    \hfill
    \subfloat[\label{fig:label_error}]{%
       \includegraphics[width=0.47\textwidth]{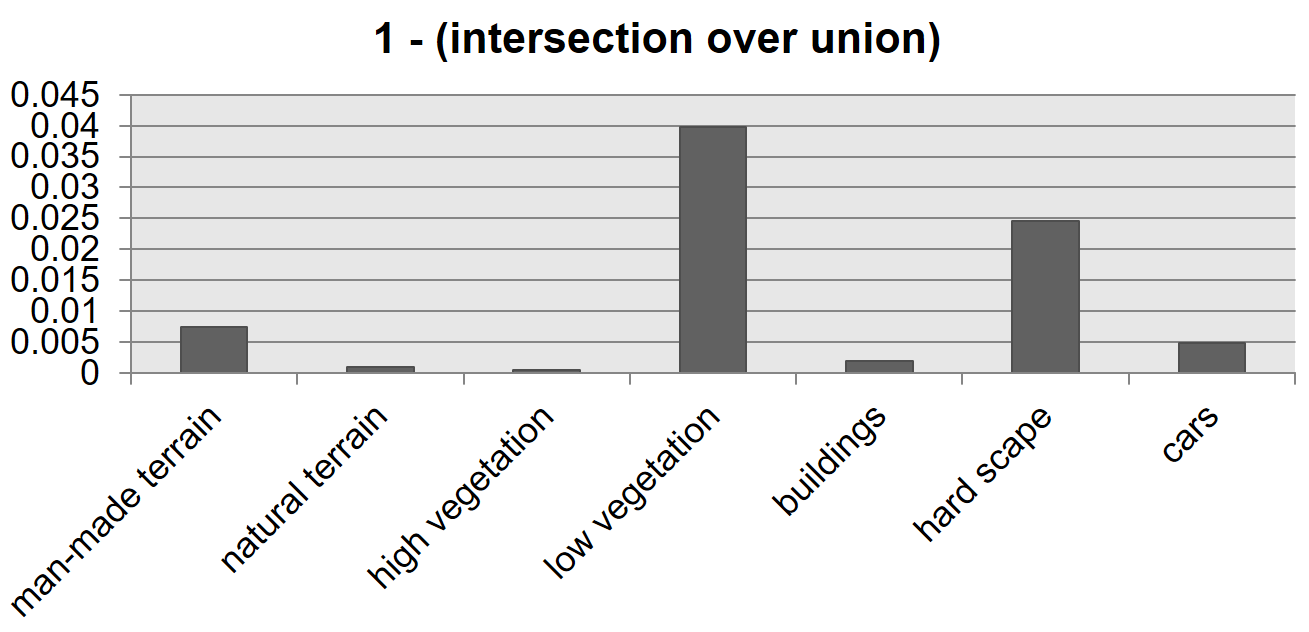}
    }
    \caption{(a) Number of points per class over all scans and (b) ground truth label errors estimated in overlapping parts of adjacent scans.}
      \label{fig:labeling}
\end{figure*}

\section{Benchmark Statistics}\label{sec:benchmark_statistics}
\vspace{-0.5em}

Class distributions in test and training set are rather similar, as
shown in Figure \ref{fig:class_distr}.  Interestingly, the class with
most samples is \textit{man-made terrain} because, out of convenience,
operators in the field tend to place the scanner on flat and paved
ground.  Recall also the quadratic decrease of point density with
distance to the scanner, such that many samples are close to the
scanner.
The largest difference between samples in test and training sets
occurs for class \textit{building}. However, this does not seem to
affect the performance of the submissions so far. Most difficult
classes, \textit{scanning artefacts} and \textit{cars}, have only few
training and test samples and a large variation of possible object
shapes. \textit{Scanning artefacts} is probably the hardest class
because the shape of artefacts mostly depends on the movement of
objects during the scanning process.
Note that, following discussions with industry professionals, class
\textit{hard scape} was designed as clutter class that contains all
sorts of man-made objects except houses, cars and ground.

In order to get an intuition of the quality of the manually acquired
labels, we also checked the label agreement among human
annotators. This provides an indicative measure on how much different
annotators agree in labeling the data, and can be viewed as an
internal check of manual labeling precision.
%One approach to perform this evaluation is to let different human annotators relabel the same scene and evaluate the different annotations.
We roughly estimate the label agreement of different human annotators
in areas where different scans of the same scene overlap. Because we
cannot rule out completely that some overlapping areas might have been
labeled by the same person (labeling was outsourced and we thus do not
know exactly who annotated what), this can only be viewed as an
indicative measure. Recall that overlaps of adjacent scans can
precisely be established via artificial markers in the scene.
Even if scan alignments would be perfect without any error, no exact
point-to-point correspondences exist between two scans, because scan
points acquired from two different locations will never exactly fall
onto the same spot. We thus have to resort to nearest neighbor search
to find point correspondences. Moreover, not all scan points have a
corresponding point in the adjacent scan. A threshold of $5$ cm on the
distance is used to ignore those points where no correspondence
exists. Once point correspondences have been estblished, it is
possible to transfer ground truth labels from one cloud to the other
and compute a confusion matrix. Note that this definition of
correspondence is not symmetric, point correspondences from cloud A in
cloud B are not equal to correspondences of cloud B in cloud A.
For each pair we calculate two intersection-over-union ($IoU_i$)
values which indicate a maximum label disagreement of $< 5\%$. No
correspondences can be found on moving objects of course, hence we
ignored category \textit{scanning artefacts} in the evaluation in
Figure \ref{fig:label_error}.

\section{Conclusion and Outlook}
\vspace{-0.5em}
\label{sec:conclusion}

The \textit{semantic3D.net} benchmark provides a large set of high
quality terrestrial laser scans with over 4 billion manually annotated
points and a standardized evaluation framework.  The data set has been
published recently and only has few submissions, yet, but we are
optimistic this will change in the future. First submissions already
show that finally CNNs are beginning to outperform more conventional
approaches, for example our covariance baseline, on large 3D laser
scans. Our goal is that submissions on this benchmark will yield
better comparisons and insights into the strengths and weaknesses of
different classification approaches for point cloud processing, and
hopefully contribute to guide research efforts in the longer term. We
hope the benchmark meets the needs of the research community and
becomes a central resource for the development of new, efficient and
accurate methods for classification in 3D space.

%\textcolor{red}{bib-entires still very long and not very consistent, should be fixed and streamlined}
%\textcolor{blue}{@Timo: Please unify bib-entries (e.g., ISPRS annals papers are often wrongly referenced)}

\section*{Acknowledgement}
\vspace{-0.5em}
This work is partially funded by the Swiss NSF project 163910, the Max Planck CLS Fellowship and the Swiss CTI project 17136.1 PFES-ES.

{
  \begin{spacing}{0.8}% tune the size by altering the parameter
    \bibliography{point_cloud}
  \end{spacing}
}

%{\small
%\bibliographystyle{ieee}
%\bibliography{point_cloud}
%}

\end{document}